\documentclass[preprint]{article}

\PassOptionsToPackage{numbers, compress}{natbib}

\usepackage{multirow}
\usepackage{multicol}
\usepackage{neurips_2024}
\usepackage{algorithm}
\usepackage{algpseudocode}
\usepackage{graphicx,subfig,blindtext}
\usepackage{newunicodechar}
\usepackage{threeparttable}

\usepackage[utf8]{inputenc} 
\usepackage[T1]{fontenc}    
\usepackage{hyperref}       
\usepackage{url}            
\usepackage{booktabs}       
\usepackage{amsfonts}       
\usepackage{nicefrac}       
\usepackage{microtype}      
\usepackage{xcolor}         
\usepackage{graphicx}
\usepackage{caption}
\usepackage{wrapfig}
\usepackage{enumitem} 
\usepackage{colortbl}
\usepackage[utf8]{inputenc}
\definecolor{darkred}{RGB}{200, 0, 0}
\definecolor{darkblue}{RGB}{0, 0, 200}

\title{Self-Evolution Fine-Tuning for Policy Optimization}

\author{
    Ruijun Chen, Jiehao Liang, Shiping Gao, Fanqi Wan, Xiaojun Quan\thanks{Corresponding author.} \\
    School of Computer Science and Engineering, Sun Yat-sen University, China \\
    \texttt{\{chenrj8,liangjh226,gaoshp,wanfq\}@mail2.sysu.edu.cn} \\ \texttt{quanxj3@mail.sysu.edu.cn} \\
}
\DeclareUnicodeCharacter{251C}{\textbar}
\DeclareUnicodeCharacter{2514}{\textbar}

\begin{document}

\maketitle

\begin{abstract}
The alignment of large language models (LLMs) is crucial not only for unlocking their potential in specific tasks but also for ensuring that responses meet human expectations and adhere to safety and ethical principles. Current alignment methodologies face considerable challenges. For instance, supervised fine-tuning (SFT) requires extensive, high-quality annotated samples, while reinforcement learning from human feedback (RLHF) is complex and often unstable. In this paper, we introduce \emph{self-evolution fine-tuning} (SEFT) for policy optimization, with the aim of eliminating the need for annotated samples while retaining the stability and efficiency of SFT. SEFT first trains an adaptive reviser to elevate low-quality responses while maintaining high-quality ones. The reviser then gradually guides the policy's optimization by fine-tuning it with enhanced responses. One of the prominent features of this method is its ability to leverage unlimited amounts of unannotated data for policy optimization through supervised fine-tuning. Our experiments on AlpacaEval 2.0 and MT-Bench demonstrate the effectiveness of SEFT. We also provide a comprehensive analysis of its advantages over existing alignment techniques.
\end{abstract}

\section{Introduction}
\label{sec:introduction}
Recent years have showcased the remarkable capabilities and performance of large language models (LLMs) across a broad range of tasks. These capabilities are attributed not only to their vast parameter sizes and the extensive text corpora used for pre-training \cite{kaplan2020scaling} but also to the critical process of aligning these models with human expectations \cite{ouyang2022training}. Such alignment is essential to ensure that the outputs of LLMs are helpful, honest, and harmless \cite{askell2021general} across various tasks and applications.

The pursuit of aligning LLMs with human preferences has led to three predominant methodologies, among others: supervised fine-tuning (SFT), reinforcement learning from human feedback (RLHF) \cite{christiano2017deep}, and offline RLHF. SFT directly fine-tunes LLMs on downstream tasks using instruction-following data, guiding the models to produce responses that match the ground truth in the dataset \cite{chung2024scaling}. RLHF offers a sophisticated approach by initially training a reward model that assigns higher rewards to responses aligning better with human preferences, followed by optimizing the policy LLM using policy-gradient methods like proximal policy optimization (PPO) \cite{schulman2017proximal}. In offline RLHF, exemplified by direct preference optimization (DPO) \cite{rafailov2024direct}, the LLM policy is directly optimized using pre-collected preference data, eliminating the necessity for a reward model. This approach aims to enhance the probability of producing \emph{chosen} responses while reducing the likelihood of \emph{rejected} ones.

Each of these methods comes with its strengths and weaknesses. SFT, while efficient, is hindered by the scarcity of high-quality human-annotated data and tends to suffer from poor adaptability to out-of-distribution samples \cite{kirk2023understanding}.
RLHF demands substantial computational overhead for training an additional reward model \cite{casper2023open} and faces optimization challenges such as inefficiency and instability.
Offline RLHF methods, which are directly optimized on preference data without the need for a reward model, tend to suffer from distribution drift issues and may lead to biased policies that favor out-of-distribution responses \cite{xu2024dpo}, thereby impairing generalization to unseen samples.

\begin{wrapfigure}{o}{0.45\textwidth}
  \vspace{-0.45cm}
  \centering
\includegraphics[width=0.45\textwidth]{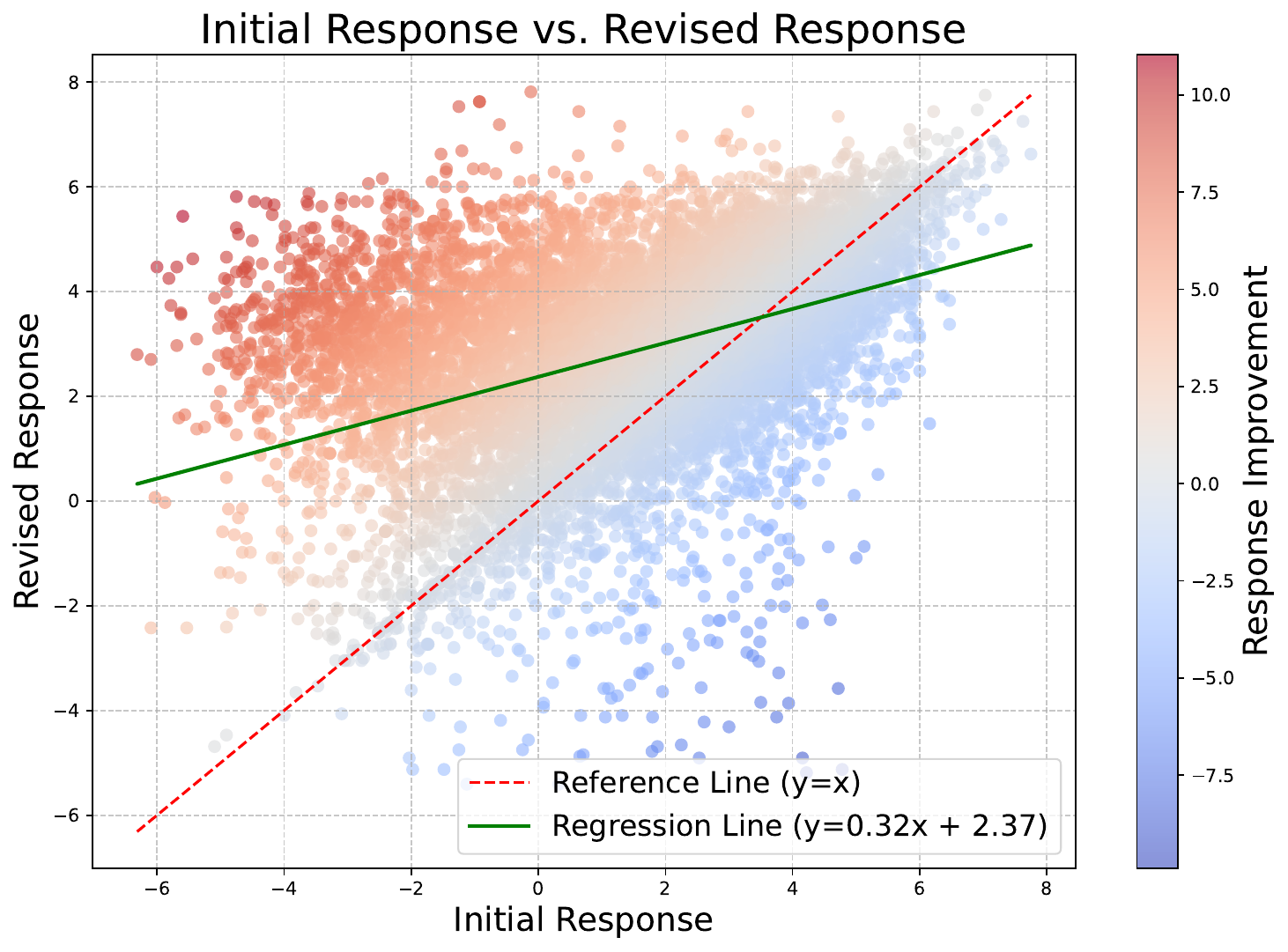}
  \caption{Reward scores of initial and revised responses on our Nectar test set \cite{starling2023}. OpenChat-3.5-7B \cite{wang2023openchat} is employed as the base model for training the reviser on Nectar's training set, and Starling-RM-7B-alpha \cite{starling2023} is used for scoring each response. Each point represents the scores of an initial response (x-axis) and the revised response (y-axis). 
  The red dashed line shows where each pair of scores is equal, and the green line shows the trend of score changes after revisions.}
  \label{fig:aligner_score}
  \vspace{-0.55cm}
\end{wrapfigure}
In response to these challenges, this paper introduces a novel \emph{self-evolution fine-tuning} (SEFT) method for policy optimization. Our approach first trains a \emph{preliminary reviser} based on preference data, which takes a prompt and the raw response of an LLM as input and outputs a higher-quality response. Such training data is widely available in various preference datasets and can be easily created from supervised fine-tuning data. However, our initial experiments suggest that the preliminary reviser tends to revise the original responses indiscriminately,
regardless of its capabilities, which 
may have a counterproductive effect on high-quality responses. As illustrated in Figure \ref{fig:aligner_score}, the reviser generally enhances low-quality responses but occasionally degrades high-quality ones.

Therefore, we continue to train an \emph{adaptive reviser} that learns to assign a revision label based on the difficulty of revising the initial response as evaluated by the preliminary reviser: \texttt{[Major Revise]} indicates a substantial revision, \texttt{[Minor Revise]} signifies a minor revision, and \texttt{[No Revise]} means that no revision is needed. The primary purpose of introducing the revision label is to guide the reviser to make revisions where feasible and refrain from attempting those beyond its capability. This ensures the overall quality of the responses produced by the reviser.

The adaptive reviser can be used to assess the quality of the policy's outputs and improve low-quality responses to high-quality ones for subsequent fine-tuning of the policy. This adaptive mechanism aligns the policy with human preferences without the need for exhaustive annotated data or complex explorations required in RLHF, and offers an innovative self-evolution fine-tuning (SEFT) strategy that balances efficiency and stability during policy optimization. The rationale behind SEFT is that it utilizes the effectiveness of pseudo-labels generated by powerful LLMs for fine-tuning. Various studies \cite{burns2023weak} have shown that pseudo-labeling enables models to be fine-tuned with high-quality synthetic labels created by robust LLMs like GPT-4. These labels approximate human annotations and allow the policy to be aligned with human-like responses at minimal cost.

To evaluate the proposed SEFT, we implemented the adaptive reviser on Nectar \cite{starling2023} using renowned LLMs of various scales, including OpenChat-3.5-7B \cite{wang2023openchat}, Qwen1.5-32B-Chat \cite{qwen}, and Yi-34B-Chat \cite{young2024yi}. We then performed policy optimization with Zephyr-7B-SFT-full \cite{tunstall2023zephyr} as the base model on UltraFeedback \cite{cui2023ultrafeedback}, where only the prompts were used. The optimized policy was evaluated on the AlpacaEval 2.0 \cite{dubois2024length} and MT-Bench \cite{zheng2024judging} benchmarks.
Experimental results demonstrate the effectiveness of SEFT and highlight its superiority over traditional alignment methods such as SFT, DPO \cite{rafailov2024direct}, and ORPO \cite{hong2024reference}. The empirical results also reveal that the proposed adaptive reviser can effectively assess the difficulty of revising responses and enhance the overall quality of model outputs.

Furthermore, experiments with additional unlabeled data from OpenHermes-2.5 \cite{OpenHermes2.5} show that incorporating more unlabeled data consistently enhances the performance of SEFT. Besides, ablation studies indicate that incrementally improving the quality of alignment data during policy optimization leads to better performance gains than directly applying high-quality alignment data. This further highlights the rationality of the proposed internal evolution and external evolution in SEFT.

\vspace{-0.2cm}
\section{Related Work}

\vspace{-0.3cm}
As large language models (LLMs) continue to advance and grow in capability, aligning them with human expectations and values becomes increasingly urgent and complex. In this section, we briefly review the mainstream methods for aligning LLMs and highlight how our approach differs.

\vspace{-0.1cm}
\subsection{Alignment Methods}
\vspace{-0.2cm}
\paragraph{SFT} Supervised fine-tuning (SFT) bridges the gap between the pre-training objective of language modeling in LLMs and the adaptation objective of making LLMs follow human instructions \cite{zhang2023instruction}. Representative SFT approaches with instruction-following data are as follows. Flan-T5  \cite{chung2024scaling} leverages Flan instruction-tuning data \cite{longpre2023flan} on a series of T5 models \cite{raffel2020exploring}, achieving improved few-shot performance. Vicuna \cite{vicuna2023} utilizes a dataset of 70K user-ChatGPT dialogues from ShareGPT\footnote{https://sharegpt.com/} and is built on the Llama-1-13B model \cite{touvron2023llama1}. It was claimed to achieve performance comparable to larger and powerful LLMs. UltraLM \cite{cui2023ultrafeedback} fine-tunes the Llama2-13b model \cite{touvron2023llama2} with UltraChat 200K instances \cite{ding2023enhancing}, once achieved the top-1 rank on the AlpacaEval leaderboard \cite{alpaca_eval}. These works illustrate the efficacy of using annotated instruction-tuning data to fine-tune LLMs for alignment.

\paragraph{RLHF} Reinforcement learning from human feedback (RLHF) has emerged as a powerful method for effectively aligning LLMs by incorporating human feedback into the learning process. This approach relies on substantial datasets of human preferences to train a reward model, which evaluates policy responses and guides the optimization process. As a pioneering work, the integration of proximal policy optimization (PPO) \cite{schulman2017proximal} for RLHF has led to notable successes in advanced models like InstructGPT \cite{ouyang2022training}. Moreover, initiatives like RLAIF \cite{lee2023rlaif} aim to address the scarcity of human preference data by generating synthetic datasets (e.g., UltraFeedback \cite{cui2023ultrafeedback} and Nectar \cite{starling2023}) using super models like GPT-4. Besides, several works try to improve RLHF by mitigating its inherent challenges. For example, APA \cite{zhu2023fine} employs a squared error loss function that incorporates estimated advantages, providing stable control over policy deviations and preventing mode collapse.

\paragraph{Offline RLHF}  Offline RLHF methods like DPO \cite{rafailov2024direct} employ the idea of contrastive learning to avoid the construction of reward models as well as the complex process of reinforcement learning. Representative works in this line also include SLiC-HF \cite{zhao2023slic}, which aligns model outputs with human preferences by incorporating two losses: calibration loss adjusts the model by increasing the likelihood of generating positive responses relative to negative ones, while regularization loss discourages the model from deviating too far from the reference model. 
IPO \cite{azar2024general}, as part of a general framework that includes RLHF and DPO, directly optimizes a pairwise preference objective with KL regularization to maintain policy alignment with a reference policy, thereby avoiding overfitting associated with traditional models that rely on pointwise reward substitution.
Furthermore, KTO \cite{ethayarajh2024kto} introduces a simple binary signal to determine the desirability of outputs, eliminating the need for expensive preference data and offering a cost-effective solution for real-world 
applications.

\paragraph{Rejection Sampling} Rejection sampling is a sample generation technique that produces candidate samples from a proposal distribution and then decides whether to accept or reject each candidate based on a certain criterion. In the context of LLM alignment, rejection sampling is often employed for efficient data augmentation. This process generates multiple responses for each prompt and then accepts or rejects these responses based on scores assigned by reward models or human evaluators. The highest-scoring responses are collected for supervised fine-tuning, a process also known as Best-of-N sampling \cite{touvron2023llama2}. Moreover, these pairs of responses consisting of high-scoring and low-scoring candidates can also be gathered as preference data for alignment training \cite{liu2023statistical,khaki2024rs}.

\subsection{Discussions}

The proposed SEFT method differs from previous approaches by focusing primarily on optimizing policy through progressively revised responses, achieved by training an adaptive reviser on existing preference data. This mechanism enables the alignment of LLMs without the need for extensive data annotations or complex explorations as required by SFT or RLHF. With the adaptive reviser, we can fine-tune the policy using unlimited amounts of unannotated instruction data. This self-evolution process ensures both efficiency and stability during policy optimization.

In the context of revising LLM outputs, Ji et al. \cite{ji2024aligner} introduced a method called Aligner, which employs a smaller LLM to refine the outputs of a primary LLM, mainly aiming to enhance the responses' usefulness and safety. Unlike Aligner, our SEFT focuses on optimizing policy models through a supervised fine-tuning process using progressively enhanced responses. Moreover, Aligner requires specific training data that involves GPT-4 for annotations, whereas SEFT utilizes existing preference or supervised fine-tuning data, reducing the need for data annotations and simplifying the process. Lastly, while Aligner enforces the revision of a response, SEFT introduces an adaptive strategy that first evaluates the difficulty of revising the response and then decides whether to revise it.

\begin{figure}[t!]
  \begin{center}
\includegraphics[width=1.00\textwidth]{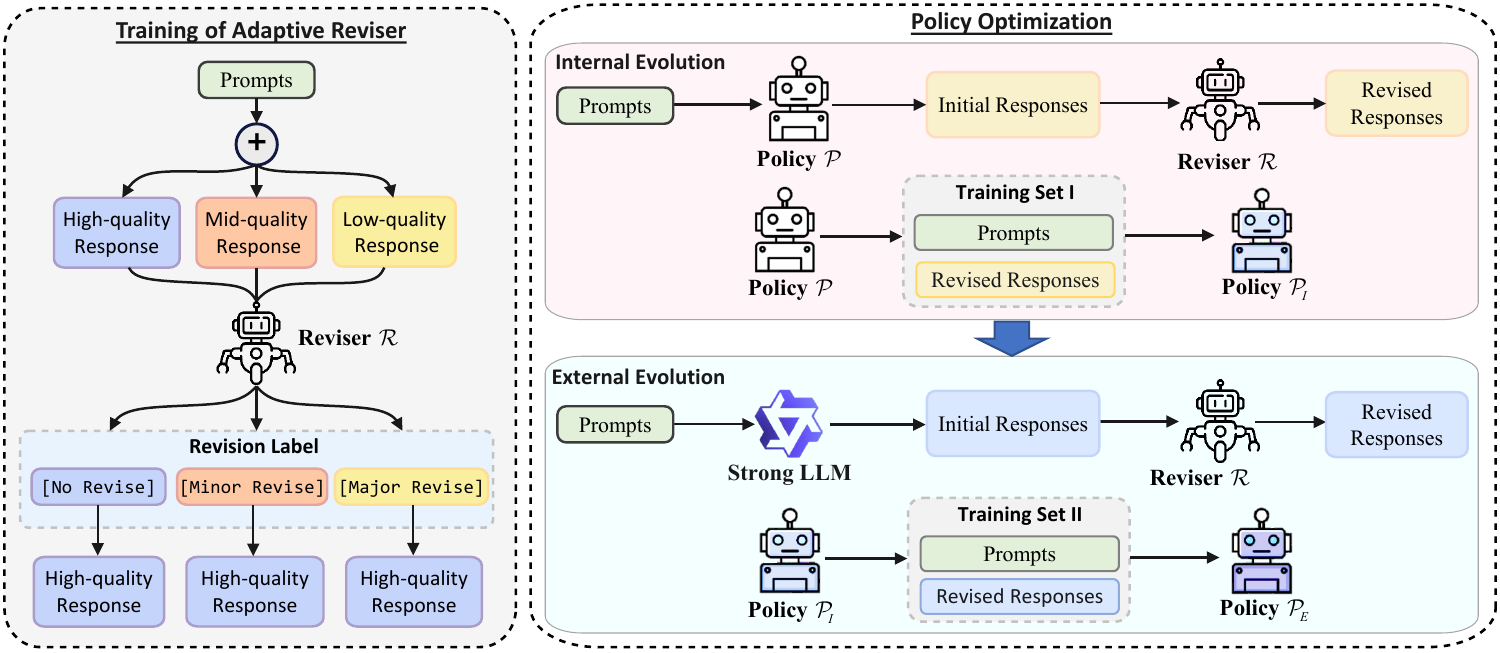}
  \end{center}
  \caption{Overview of SEFT. The reviser takes prompts and initial responses of varying quality as input, assesses the difficulty of revising these responses, and assigns appropriate revision labels to generate overall high-quality responses. During policy optimization, the policy first undergoes internal evolution: the reviser revises the responses generated by the policy and uses them to fine-tune the policy. Then, the policy undergoes external evolution with a stronger model, progressively enhancing the quality of alignment data and guiding the policy toward generating better responses.}
  \label{fig:overview}
\end{figure}
\section{SEFT: Self-Evolution Fine-Tuning}
\label{SEFT}

Self-Evolution Fine-Tuning (SEFT) aims to provide a robust and efficient solution for policy optimization. To achieve this, as shown in Figure \ref{fig:overview}, we first train an adaptive reviser that evaluates the initial responses generated by the policy and makes appropriate revisions to enhance the overall quality. Then, the policy is fine-tuned using these enhanced responses as pseudo-labels. This self-evolution mechanism allows us to optimize the policy without exhaustive annotated data or complex explorations while maintaining the stability and efficiency of supervised fine-tuning (SFT). Below, we first provide an overview of the SEFT method in Section \ref{subsec:overview}. We then give a detailed explanation of the training process for the adaptive reviser in Section \ref{Reviser}, followed by the policy optimization using the reviser in Section \ref{alignment}.

\subsection{Overview}
\label{subsec:overview}

\begin{figure}[t!]
  \begin{center}
\includegraphics[width=1.00\textwidth]{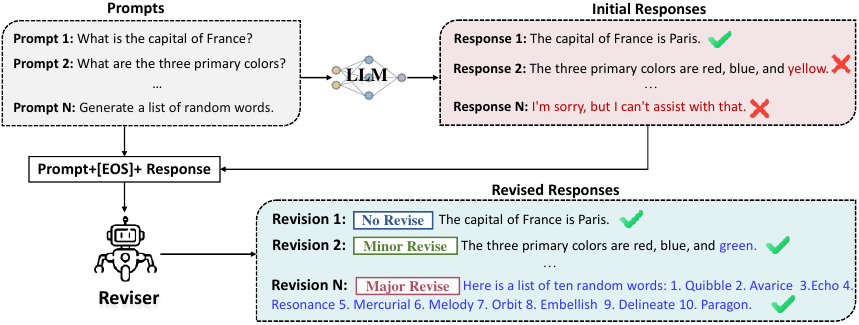}
  \end{center}
  \caption{Illustrative training examples for the adaptive reviser. The objective of the adaptive reviser is to make revisions where feasible and avoid attempting those beyond its capabilities.
  }
  \label{fig:reviser_example}
\end{figure}

As depicted in Figure \ref{fig:reviser_example}, the proposed SEFT begins with training an adaptive reviser  $\mathcal{R}$ using prompts (instructions) and responses of varying quality. For each prompt, a pair of responses is provided, one denoting a low-quality response and the other a high-quality response. Such training data is broadly available in various preference datasets \cite{cui2023ultrafeedback} and can also be easily created from supervised fine-tuning data. In preference datasets, for example, the low-quality response and high-quality response correspond to \emph{reject} and \emph{chosen} responses, respectively. The training of the reviser starts from a strong base model $\mathcal{M}$ and involves an initial warm-up phase to create a preliminary reviser, followed by adaptive training to continually refine the reviser. This enables the reviser to adaptively revise a response based on the difficulty of the revision and minimize misrevisions.

We then apply the reviser twice for policy optimization on unlabeled prompts. First, initial responses for these prompts are sampled from the policy 
\( \mathcal{P}\), and the reviser is applied to refine these responses. These revised responses serve as pseudo-labels to fine-tune the policy. Next, the base model 
\( \mathcal{M}\) is employed to generate another set of initial responses for these prompts, which are also refined by the reviser \( \mathcal{R}\) to enhance their quality. These enhanced responses are then used to fine-tune the policy once more. This two-phase approach forms an internal-external evolution path, which allows 
us to first fine-tune \( \mathcal{P}\) within its own response space before expanding to a more challenging response space, promoting progressive improvements. This process also follows the idea of curriculum learning \citep{wang2022survey}.

One might naturally expect an iterative application of the reviser to optimize the policy, but this proves to be infeasible. This is because the reviser does not perform iterative training but directly refines initial responses into final responses, preventing gradual improvement over multiple iterations. Our experiments also confirm that the policy does not benefit from the second round of optimization.

\subsection{Reviser Training}
\label{Reviser}

Given a revision dataset \( \mathcal{D}^r = \{X^r, {Y^l}, {Y^h}\}\), where $X^r=\{X^r_i\}|_{i=1}^N$ stands for a set of prompts, $Y^l=\{Y_i^l\}|_{i=1}^N$ denotes the original low-quality responses for these prompts, and $Y^h=\{Y_i^h\}|_{i=1}^N$ are the corresponding responses of higher quality. For simplicity, this work leverages an existing preference dataset for \( \mathcal{D}^r \), where \emph{rejected} responses correspond to $Y^l$ and \emph{chosen} responses correspond to $Y^h$. We partition \( \mathcal{D}^r \) into two distinct splits \(\mathcal{D}^r = \{\mathcal{D}_1^r, \mathcal{D}_2^r\}\) for warm-up training and adaptive training, respectively. The overall training objective of the reviser \( \mathcal{R} \) is to assess the level of difficulty in revising \( {Y}^l \) to \( {Y}^h \) and implement adaptive revisions.

\paragraph{Warm-up Training}
\label{preliminary reviser}

We begin with warm-up training on \( \mathcal{D}_1^r \) to obtain a preliminary reviser \( \hat{\mathcal{R}} \), aiming to transform the low-quality responses into high-quality responses. This process can be represented as $\hat{\mathcal{R}}: X^r\times Y^l \rightarrow {Y}^h$. Specifically, the training objective is to minimize the negative log-likelihood (NLL) of high-quality responses \( {Y}^h \) given prompts \( {X^r} \) and low-quality responses \( {Y}^l \):
\begin{equation}
\label{eq:stage_one_training}
    \mathcal{L}_r = -\mathbb{E}_{(X_i^r,Y_i^l,Y_i^h) \sim \mathcal{D}_1^r}\left[\mathrm{log}P_{\theta}(Y_i^h|X_i^r, Y_i^l)\right], 
\end{equation}
where \( \theta \) refers to the parameters of the reviser \( \hat{\mathcal{R}} \) initialized from the base model \( \mathcal{M} \).
Note that this training process enables the reviser to utilize information from both the prompts in $X^r$ and the initial responses in $Y^l$ to generate the final responses $Y^h$.

\paragraph{Adaptive Training}

The above preliminary reviser \( \hat{\mathcal{R}} \) tends to revise the original responses indiscriminately, regardless of its capabilities, which may lead to the deterioration of high-quality responses, as illustrated in Figure \ref{fig:aligner_score}.
Ideally, we want the reviser to make revisions where feasible and avoid attempting those beyond its capability. To achieve this, we define three labels based on the difficulty of revising the initial responses as evaluated by the preliminary reviser \( \hat{\mathcal{R}} \): \texttt{[Major Revise]}, \texttt{[Minor Revise]}, and \texttt{[No Revise]}. \texttt{[Major Revise]} indicates a substantial revision, \texttt{[Minor Revise]} signifies a minor revision, and \texttt{[No Revise]} means no revision is needed.

Since the original training dataset lacks revision labels, an intuitive approach is to use human labor for annotations. However, this method is not only expensive but may also introduce potential biases from human annotators \cite{sun2024principle}. To automatically obtain revision labels, we apply the preliminary reviser  \( \hat{\mathcal{R}} \) on \( \mathcal{D}_2^r \) and evaluate its revisions as follows. For each sample in \(\mathcal{D}_2^r\) consisting of a prompt \( X_i^r \) and the original response \( {Y}_i^l \), we first apply \( \hat{\mathcal{R}} \) to revise \( {Y}_i^l \). We then use an off-the-shelf critic model \(\mathcal{C}\) to assess the revised response \( \hat{Y}_i^l\) against the original response $Y_i^l$, assigning a benefit score \( s_i \) for the revision to measure the extent of improvement achieved by the preliminary reviser. We compare the reward \( s_i \) with two thresholds, $\delta_l$ and $\delta_h$, to assign an appropriate revision label \( r_i \) for the prompt \( X_i^r \). If the reward \( s_i \) is higher than $\delta_h$, it indicates that the current sample is easy for the reviser \( \hat{\mathcal{R}} \) to revise, allowing for a major revision of \(Y_i^l\). If the reward \( s_i \) is between $\delta_l$ and $\delta_h$, it suggests that the reviser can improve the original response moderately, so a minor revision is appropriate. Otherwise, if the reward \( s_i \) is lower than $\delta_l$, it indicates that the reviser struggles to improve the response, and the original response should remain unchanged. Moreover, considering that the reviser's ability may improve with further training on \(\mathcal{D}_2^r\), we further use a probability $p$ to control the proportion of revision labels for samples in the latter two categories.

After assigning each prompt a revision label, we use the updated \(\mathcal{D}_2^r\) to continually train the preliminary reviser using the NLL objective:
\begin{equation}
\label{eq:stage_two_training}
    \mathcal{L} = -\mathbb{E}_{(X_i^r,Y_i^l,r_i,Y_i^h) \sim \mathcal{D}_2^r}\left[\mathrm{log}P_{\theta}(r_i, Y_i^h|X_i^r, Y_i^l)\right].
\end{equation}
Unlike warm-up training, the adaptive reviser learns to predict the revised response as well as the revision label based on the difficulty of revising the initial response, thus implementing adaptive revisions. 
The overall training process of the adaptive reviser is illustrated in Algorithm \ref{alg: reviser}.

\begin{algorithm}[t]
\caption{Reviser Training}\label{alg: reviser}
\begin{algorithmic}[1]
\Require base model \( \mathcal{M} \), dataset \( \mathcal{D}^r = \{\mathcal{D}_1^r, \mathcal{D}_2^r\} = \{X_i^r, Y_i^l, Y_i^h\}\), critic model \(\mathcal{C}\), hyperparameters $\delta_l$, $\delta_h$, and $p$ for revision label classification.

\State Fine-tune \( \mathcal{M} \) on dataset \(\mathcal{D}_1^r \) according to Eq.(\ref{eq:stage_one_training}) to obtain the preliminary reviser \( \hat{\mathcal{R}} \).
\For {$ (X_i^r, Y_i^l, Y_i^h) $ in $\mathcal{D}_2^r$}
\State Use \( \hat{\mathcal{R}} \) to generate the revised response \(\hat{Y}_i^l\).
\State Use \(\mathcal{C}\) to score \( Y_i^l \)  and \(\hat{Y}_i^l\), and calculate the benefit score \( s_i = \texttt{Score} (\hat{Y}_i^l) - \texttt{Score} (Y_i^l)\).
\If{\( s_i \) > $\delta_h$}
\State Revision label \( r_i \) = [\texttt{Major Revise}].
\ElsIf{ $\delta_h$ > \( s_i \) > $\delta_l$}
\State  With probability $p$, revision label \( r_i \) = [\texttt{Minor Revise}].
\State  With probability $1-p$, revision label \( r_i \) = [\texttt{No Revise}].
\ElsIf{ \( s_i \) < $\delta_l$}
\State  With probability $p$, revision label \( r_i \) = [\texttt{No Revise}]
\State  With probability $1-p$, revision label \( r_i \) = [\texttt{Minor Revise}].
\EndIf 
\State Update \( \mathcal{D}_2^r \) with \(({X_i^r}, Y_i^l, r_i, Y_i^h)\) 
\EndFor
\State Continually fine-tune \( \hat{\mathcal{R}} \) on dataset \(\mathcal{D}_2^r \) according to Eq.(\ref{eq:stage_two_training}) to obtain the adaptive reviser \( \mathcal{R} \).
\end{algorithmic}
\end{algorithm}

\subsection{Alignment}
\label{alignment}

In this section, we elaborate on the details of the proposed SEFT for policy optimization. The SEFT process can be generally defined as follows:
\begin{equation}
    \mathcal{P} = \mathbb{SEFT}(X^p, \mathcal{P}, \mathcal{G}, \mathcal{R}),
    \label{eq:seft_general}
\end{equation}
where $X^p = \{X_j^p\}|_{j=1}^M$ denotes the set of unannotated prompts for the optimization of policy $\mathcal{P}$, $\mathcal{G}$ is the generator used to generate an initial response $\hat{Y}_j^p$ for each prompt $X_j^p$, and $\mathcal{R}$ is the adaptive reviser used to revise the initial response to obtain an enhanced response $Y_j^p$. The prompts $X^p$ and the enhanced responses $Y^p = \{Y_j^p\}|_{j=1}^M$ will be used to fine-tune the policy $\mathcal{P}$ as follows:
\begin{equation}
\label{eq:policy_opt}
   \mathcal{P} = \arg \min_{\phi} - \mathbb{E}_{(X_j^p, Y_j^p)} \left[\mathrm{log}P_\phi(Y_j^p|X_j^p)\right],
   \label{eq:seft_finetune}
\end{equation}
where \( \phi \) refers to the parameters of the policy, typically initialized from an SFT model \cite{rafailov2024direct}.

The policy optimization with SEFT involves both internal and external evolution:
\vspace{-0.3cm}
\paragraph{Internal Evolution} In the internal evolution, the policy \(\mathcal{P}\) is optimized using its own generated responses, which are revised by the adaptive reviser \(\mathcal{R}\). This phase focuses on improving the policy within its own response space. Acting as the generator \(\mathcal{G}\) in Equation \ref{eq:seft_general}, the policy \(\mathcal{P}\) first generates initial responses for the prompts \(X^p\). These initial responses are then refined by the reviser \(\mathcal{R}\) to produce a set of high-quality responses. Finally, the policy \(\mathcal{P}\) is fine-tuned using the prompts and the revised responses as outlined in Equation \ref{eq:seft_finetune}, resulting in an updated policy $\mathcal{P}_I$.

\paragraph{External Evolution} In the external evolution phase, the policy \(\mathcal{P}_I\) is further fine-tuned using revised responses from an external robust generator, which is the base model $\mathcal{M}$ of the reviser in this study. This phase aims to enhance the policy's capabilities by exposing it to a higher-quality and more challenging response space, thus facilitating more substantial improvements. Similar to the internal evolution phase, the prompts \(X^p\) and the revised responses generated by the adaptive reviser from the initial responses of the external generator $\mathcal{M}$ will be used to further fine-tune \(\mathcal{P}_I\) according to Equation \ref{eq:seft_finetune}, resulting in the final policy \(\mathcal{P}_E\).

Note that while the external evolution phase ensures that the policy benefits from superior examples, the integration of both phases enables the policy model to progressively enhance its performance.

\section{Experiments}
\label{experiment}

We conduct extensive experiments to evaluate the effectiveness of our proposed SEFT, specifically examining the reviser's ability to enhance the quality of initial responses and its impact on policy optimization. Furthermore, we conduct analyses to explore the benefits of integrating additional unlabeled data and the effects of progressive policy optimization via internal and external evolution.

\subsection{Experimental Setup}

\paragraph{Training Datasets}
We employ Nectar \cite{starling2023} for training the reviser. Nectar is a high-quality dataset comprising 183K diverse dialogues, each prompt containing seven responses generated from various models and ranked by GPT-4. To prevent data contamination, prompts appearing in subsequent evaluation benchmarks were filtered out. Then, we set aside 1.8K samples as the test set for reviser evaluation. The remaining data is divided in a 3:7 ratio, with one partition used for training the preliminary reviser and the other for the adaptive reviser described in Section \ref{Reviser}. In the Nectar dataset, Rank 0 represents the highest-quality response, whereas Rank 6 represents the lowest-quality response. Our initial responses are constructed by randomly selecting from responses ranked 1 to 6, with responses ranked 0 representing the final high-quality responses. This ensures the training data covers a diverse range of initial responses, enabling the reviser to acquire versatile capabilities.

For policy optimization, we utilize UltraFeedback\footnote{\url{https://huggingface.co/datasets/openbmb/UltraFeedback}} \cite{cui2023ultrafeedback}, which comprises 64K prompts collected from diverse sources. To validate that integrating more unannotated data consistently enhances the performance of SEFT, we additionally sample subsets of 30K and 60K prompts respectively from OpenHermes-2.5\footnote{\url{https://huggingface.co/datasets/teknium/OpenHermes-2.5}} \cite{OpenHermes2.5} as supplementary training data.

\paragraph{Training Details}
During training the reviser model, we first explore implementing it with OpenChat-3.5-7B \cite{wang2023openchat}. Then, we compare the performance with different base models, including OpenChat-3.5-7B \cite{wang2023openchat}, Qwen1.5-32B-Chat \cite{qwen}, and Yi-34B-Chat \cite{young2024yi}. Striving for a balance between efficiency and performance, we ultimately opt for Qwen1.5-32B-Chat to implement the reviser. The reviser is trained on eight NVIDIA A800 GPUs using DeepSpeed ZeRO 3. The batch size is set to 128, and the learning rate scheduler follows the cosine type, with a warmup ratio of 0.03 and a learning rate of 2e-5. The model undergoes training for 3 epochs, with a maximum length of 2,048 tokens. To obtain the revision labels, we utilize Starling-RM-7B-alpha \cite{starling2023} as the critic model \(\mathcal{C}\). The hyperparameters \(\delta_l\), \(\delta_h\), and \(p\) are set to 0.3, 1.0, and 0.8, respectively. The adaptive reviser is initialized using the preliminary reviser with identical hyperparameter settings and then trained continuously.

For the training of the policy model, we opt for Zephyr-7B-SFT-full \cite{tunstall2023zephyr} as the backbone. Zephyr-7B-SFT-full and Qwen1.5-32B-Chat are respectively employed as the generators for the internal evolution and external evolution phases, as mentioned in Section \ref{alignment}. Similar to the reviser, training is conducted using DeepSpeed ZeRO 3 on eight NVIDIA A800 GPUs. The total batch size is set to 128, with a cosine learning rate scheduler, a warmup ratio of 0.1, and a learning rate of 5e-7. The model is trained for 3 epochs for each evolution, with a maximum length of 2,048 tokens.

\paragraph{Evaluation}
To assess the reviser's effectiveness in enhancing response quality, we adopt methodologies from prior research \cite{kirk2023understanding,ji2024beavertails} and utilize established reward models to evaluate both the initial and revised responses across the Nectar test set. Specifically, we employ four robust reward models: FsfairX-LLaMA3-RM-v0.1 \cite{dong2023raft}, Eurus-RM-7B \cite{yuan2024advancing}, Starling-RM-7B-alpha \cite{starling2023}, and RM-Mistral-7B \cite{xiong2023gibbs}. The assessments generated by these reward models are then aggregated to determine the reviser's performance and overall success rate.

For the policy evaluation, we employ AlpacaEval 2.0 \cite{dubois2024length} and MT-Bench \cite{zheng2024judging}. AlpacaEval 2.0 includes 805 instructions and assesses the model's success rate against GPT-4 for each instruction, using an evaluator based on GPT-4. MT-Bench \cite{zheng2024judging} consists of 80 multi-turn dialogues spanning eight domains, with GPT-4 rating the model's responses on a scale from 1 to 10.

\paragraph{Baselines} We first compare the performance of our adaptive reviser with the following methods: (i) \textbf{Aligner} \cite{ji2024aligner} primarily aims to enhance the usefulness and safety of a primary LLM by training a smaller LLM to refine the outputs. For a fair comparison, we implement Aligner on the same training set as our reviser. (ii) \textbf{Label reviser} is similar to our adaptive reviser but assigns revision labels based on the disparity in rank between the initial response and the target response in the Nectar training set, whereas our revision labels are determined by the performance of the preliminary reviser. For more details, refer to Appendix \ref{appendix:implementation}. (iii) \textbf{Preliminary reviser}, as described in Section \ref{Reviser}, is trained using only 30\% of the training data and tries to revise all low-quality initial responses. (iv) \textbf{Original response} denotes the original input responses to a reviser, which serve as a naive baseline.

Next, we compare the proposed SEFT with other policy optimization methods: (i) \textbf{SFT} directly optimizes the policy model using prompts and \emph{chosen} responses from the UltraFeedback dataset. (ii) \textbf{DPO} \cite{rafailov2024direct} optimizes the policy model by applying a reward modeling objective to preference data. (iii) \textbf{ORPO} \cite{hong2024reference} presents a reference model-free method that integrates the odds ratio of the \emph{chosen} response over the \emph{rejected} response into the SFT objective function. This integration serves to penalize the probability of generating the rejected response directly.

\begin{figure}[t]
  \begin{center}
 \includegraphics[width=0.999\textwidth]{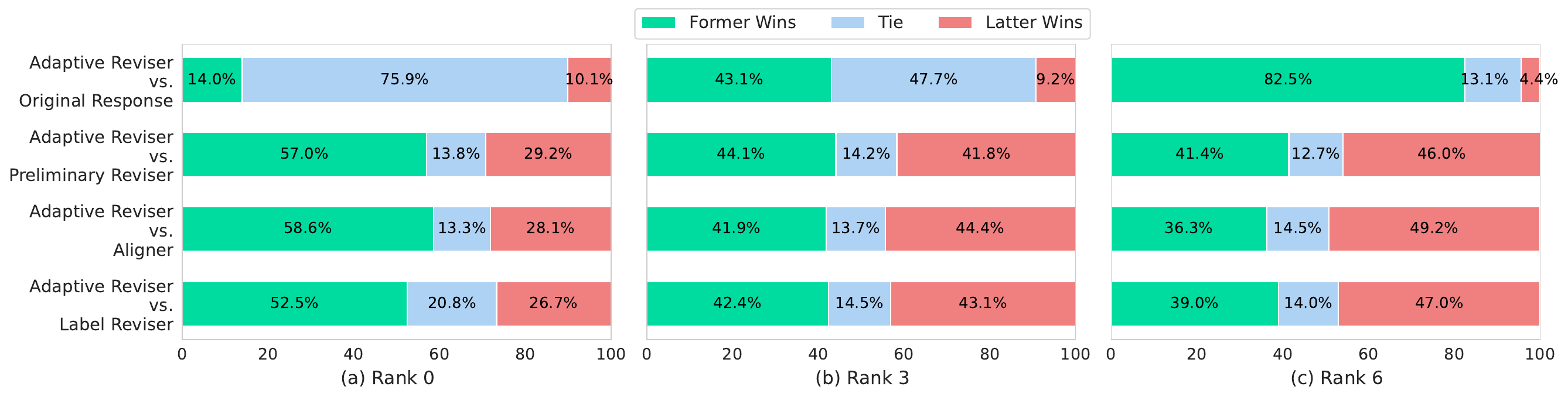}
  \end{center}
  \caption{Performance comparison of the adaptive reviser and baseline methods on the Nectar test set. Due to the limit of space, we only present the results for responses ranked 0, 3, and 6.}
  \label{fig:pre_reviser_compare}
\end{figure}
\subsection{Results of Reviser Evaluation}

In this section, we present the evaluation results of our adaptive reviser. We first compare its performance against several baseline methods. Figure \ref{fig:pre_reviser_compare} depicts the win rates of pairwise comparisons between the responses generated by our adaptive reviser and those by the baselines.  Due to the limit of space, we only present the results for responses ranked 0, 3, and 6 on the Nectar test set. More results for other ranks are provided in Appendix \ref{reviser_performence}. 

We make four key observations from the results. First, the proposed adaptive reviser outperforms the baseline methods in revising original responses across ranks 0 to 6. This is especially true for high-quality responses, such as those at Rank 0. The adaptive reviser’s success can be attributed to its adaptive strategy, which selectively revises responses it can effectively improve while avoiding those it cannot. Second, for high-quality responses, the adaptive reviser's improvements over the original responses are modest. This is because high-quality responses already approach an optimal state, leaving limited room for enhancement by the reviser. Conversely, the adaptive reviser shows significant improvement for low-quality responses. Fourth,  although the adaptive reviser generally excels, it performs slightly worse than the baselines for some low-quality responses (e.g., Rank 6). This is because the reviser often applies the \texttt{[No Revise]} label to certain instances, choosing not to revise every low-quality response and therefore impacting the overall performance. More details about the distribution of revision labels can be found in Appendix \ref{reviser_performence}.

\begin{figure}[t!]
  \begin{center}
\includegraphics[width=0.95\textwidth]{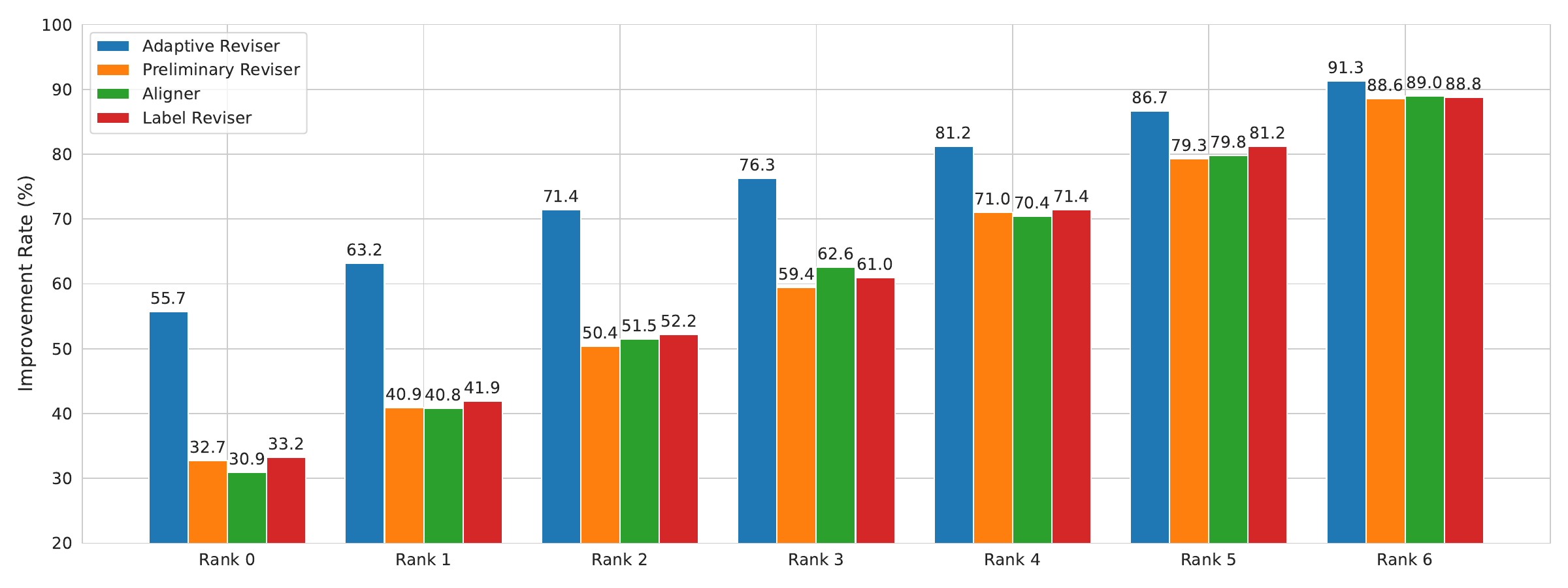}
  \caption{Improvement rate of different revisers on the Nectar test set. The metric is defined as the proportion of revised instances showing enhanced quality at each rank. Note that instances that stay unchanged by revisers are excluded from this calculation.}
  \label{fig:pre-win-rate}
  \vspace{-0.3cm}
\end{center}
\end{figure}

\begin{figure}[t]
  \begin{center}
\includegraphics[width=0.95\textwidth]{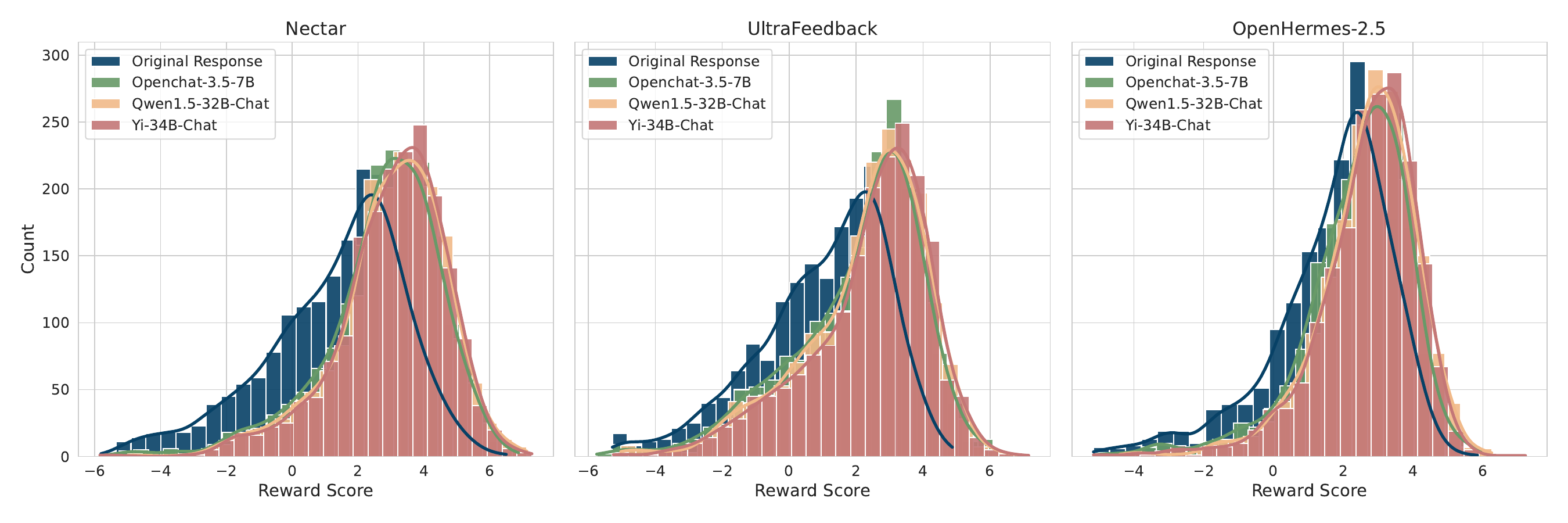}
  \end{center}
  \caption{
  Reward distributions of original responses generated by Zephyr-7B-SFT-full and revised responses produced by adaptive revisers based on various models, including Openchat-3.5-7B, Yi-34B-Chat, and Qwen1.5-32B-Chat. This comparison is conducted across three test sets: Nectar, UltraFeedback, and OpenHermes-2.5. The x-axis denotes the reward scores generated by Starling-RM-7B-alpha, while the y-axis indicates the number of instances that fall within each reward range.} 
  \label{fig:reward-transfer}
  \vspace{-0.3cm}
\end{figure}

To further illustrate the effectiveness of the adaptive reviser in enhancing response quality without compromising the integrity of high-quality responses, we present the \emph{improvement rate}. This metric is defined as the proportion of revised instances that show enhanced quality at each rank, achieved by various revisers on the Nectar test set. As shown in Figure \ref{fig:pre-win-rate}, the adaptive reviser consistently outperforms the baseline revisers across all ranks. Specifically, for responses of lower initial quality (e.g., Ranks 5 and 6), all methods exhibit high rates of improvement. However, as the quality of the initial responses increases, the improvement rates of the baseline revisers decline sharply. This suggests that these revisers may over-revise, degrading the quality of already strong responses. In contrast, the adaptive reviser maintains robust improvement rates even for high-quality responses, indicating its ability to avoid unnecessary revisions that could degrade their quality. For example, at Rank 0, the adaptive reviser achieves an improvement rate of 55.7\%, significantly higher than that of the other revisers. This demonstrates that the adaptive reviser not only facilitates beneficial revisions but also strategically avoids making unnecessary changes that might compromise response quality.

Additionally, we analyze the reward distributions of the original responses generated by Zephyr-7B-SFT-full with those of the revised responses produced by adaptive revisers based on various models, including Openchat-3.5-7B, Yi-34B-Chat, and Qwen1.5-32B-Chat. This comparison is conducted across three test sets: Nectar, UltraFeedback, and OpenHermes-2.5. The results in Figure \ref{fig:reward-transfer} reveal a consistent improvement in the reward distributions for revised responses across these test sets. Furthermore, the magnitude of this shift tends to correlate with the scale of the reviser models, indicating that larger models are more effective at enhancing response quality.

\begin{table*}[t!]
    \centering
    \resizebox{0.95\textwidth}{!}{
    \begin{tabular}{lllccc}
        \toprule
        \multirow{2}{*}{\textbf{Model}} &  ~ & \multirow{2}{*}{\textbf{Stage}} & \multicolumn{2}{c}{\textbf{AlpacaEval 2.0}} & \textbf{MT-Bench} \\
         & & & \textbf{LC Win Rate} & \textbf{Win Rate} & \textbf{Score} \\
        \midrule
        \multicolumn{2}{l}{GPT-4 0613$^\dagger$}  & - & 30.2\%  & 15.8\% &  9.18 \\
        \multicolumn{2}{l}{GPT-3.5 Turbo 0613$^\dagger$}  & - & 22.7\%  & 14.1\% &  8.39 \\
        \midrule
        \multicolumn{2}{l}{Zephyr-7B-SFT-full$^*$} &  SFT & 6.4\% & 4.4\% &  6.33 \\
        \multicolumn{2}{l}{Zephyr-7B-SFT-full-SFT$^*$} & SFT + SFT & 8.7\% & 6.7\% &  6.99 \\
        \multicolumn{2}{l}{Zephyr-7B-$\beta$$^\dagger$} &  SFT + DPO & 13.2 \% &11.0\%  &  7.34  \\
        \multicolumn{2}{l}{Mistral-ORPO-$\beta$$^\dagger$} &  ORPO & 14.7\%  & 12.6\% &  7.32 \\
        \midrule
        \rowcolor[gray]{.93} Zephyr-7B-SEFT (ours) &&&&& \\
        \,\,\,\, UltraFeedback & ~&   SFT + SEFT & 15.6\% & 11.8\% &  7.32 \\
        \,\,\,\,\,\,\,\, +30K additional data & &  SFT + SEFT & 15.2\% & 12.0\% &  7.35 \\
         \,\,\,\,\,\,\,\, +60K additional data & &  SFT + SEFT & \textbf{16.6\%} & \textbf{13.7\%} &  \textbf{7.47} \\
        \bottomrule
    \end{tabular}
    }
    \caption{Results of policy evaluation concerning instruction-following abilities, achieved by different policy optimization methods on MT-Bench and AlpacaEval 2.0. A dash ``-'' signifies results not publicly available, ``$^\dagger$'' denotes results from the leaderboard, and ``$^*$'' indicates results from our reproduction. ``Zephyr-7B-SFT-full-SFT'' refers to the further fine-tuned Zephyr-7B-SFT-full with \emph{chosen} samples from UltraFeedback. The notations ``+30K additional data'' and ``+60K additional data'' denote the inclusion of additional 30K and 60K unannotated prompts, respectively.}
    \label{tab:models}
    \vspace{-0.3cm}
\end{table*}

\vspace{-0.1cm}
\subsection{Results of Policy Evaluation}

In this subsection, we present the performance of policy models optimized with different alignment methods on AlpacaEval 2.0 and MT-Bench. As shown in Table \ref{tab:models}, the proposed SEFT demonstrates superior performance to the DPO (Zephyr-7B-$\beta$) and ORPO (Mistral-ORPO-$\beta$) methods, 
\begin{wrapfigure}{o}{0.45\textwidth}
  \vspace{-0.2cm}
  \centering
\includegraphics[width=0.95\linewidth]{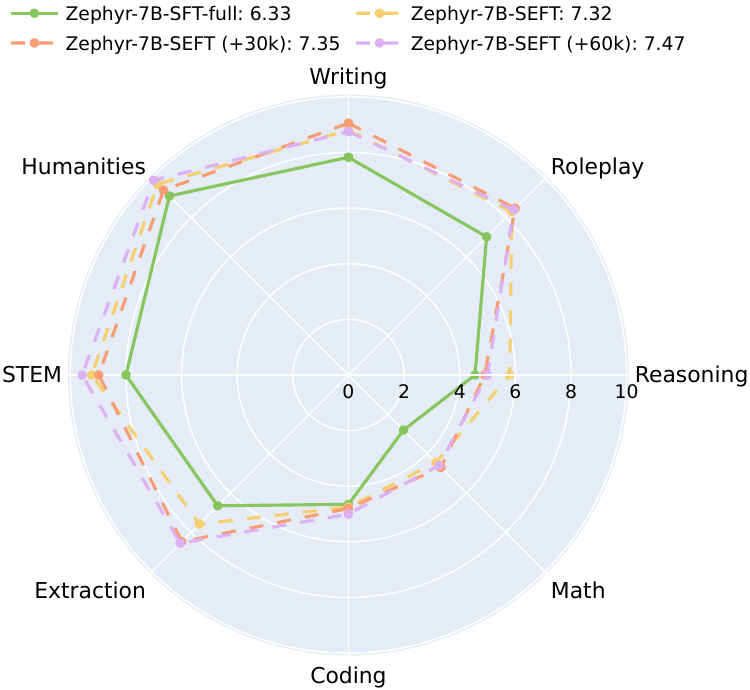}
  \caption{Comparison between Zephyr-7B-SFT-full and Zephyr-7B-SEFT on MT-Bench with varying amounts of additional data.}
  \label{fig:mt-bench}
  \vspace{-0.4cm}
\end{wrapfigure}
utilizing solely unannotated prompts from UltraFeedback for policy optimization. Specifically, the Zephyr-7B-SEFT model achieves an LC Win Rate of 15.6\%, a Win Rate of 11.8\%, and an MT-Bench Score of 7.32. These results are competitive with other methods that use supervised prompts, such as Zephyr-7B-$\beta$ (13.2\%, 11.0\%, 7.34) and Mistral-ORPO-$\beta$ (14.7\%, 12.6\%, 7.32).

Moreover, when additional unannotated prompts 
are incorporated, the performance of SEFT further improves, highlighting its effectiveness and scalability with unannotated samples. 
For example, when supplemented with an additional 60K unannotated prompts, Zephyr-7B-SEFT achieves remarkable performance, including the highest MT-Bench Score of 7.47, an LC Win Rate of 16.6\%, and a Win Rate of 13.7\%. Figure \ref{fig:mt-bench} further confirms this trend,  showing consistent improvements in Zephyr-7B-SEFT's scores across various categories on MT-Bench as the volume of additional data increases.  Specifically, the model with the 60K additional prompts demonstrates significant enhancements in categories such as \emph{Extraction} and \emph{STEM}, highlighting its improved capability to manage complex and diverse language tasks. The results clearly indicate that as more unannotated data is integrated, the model's performance not only becomes more robust across all categories but also significantly outperforms the baseline Zephyr-7B-SFT-full model. These findings emphasize the effectiveness of SEFT in leveraging unannotated data to enhance response quality while maintaining stability and efficiency in policy optimization.

\subsection{Ablation Studies}

We conduct comprehensive ablation studies for the proposed SEFT method on the MT-Bench dataset. These studies focus on the impact of each SEFT component, particularly emphasizing the 
\begin{wraptable}{r}{0.5\textwidth}
    \vspace{-0cm}
    \centering
    \resizebox{0.5\textwidth}{!}{
    \begin{tabular}{llc|c}
    \toprule
        \multirow{2}{*}{Model} & ~ & Extra & \multirow{2}{*}{MT-Bench} \\ 
        ~ & ~ &  data & ~ \\
        \midrule
        Zephyr-7B-SFT-full & ~ & - & 6.33 \\ 
        \midrule
        + SFT-Chosen & ~ & \multirow{5}{*}{-} & 6.99 \\ 
        + SFT-External Generator & ~ & ~ & 7.06 \\ 
        + External Evolution & ~ & ~ & 7.11 \\ 
        + Internal Evolution & ~ & ~ & 6.89 \\ 
         \quad\quad + External Evolution & ~ & & 7.32 \\ 
        \midrule
        + External Evolution & ~ & \multirow{3}{*}{30K} & 7.12 \\ 
        + Internal Evolution & ~ & ~ & 6.83 \\ 
        \quad\quad + External Evolution & ~ &  & 7.35 \\ 
        \midrule
        + External Evolution & ~ & \multirow{3}{*}{60K} &7.14 \\ 
        + Internal Evolution & ~ & ~ & 6.83 \\ 
        \quad\quad + External Evolution & ~ &  & 7.47 \\ 
        \bottomrule
    \end{tabular}
    }
    \caption{Results of ablation studies for SEFT on the MT-Bench dataset.
    ``SFT-Chosen'' refers to using the \emph{chosen} responses from UltralFeedback as ground truth to fine-tune the policy model, and ``SFT-External Generator'' indicates using the responses generated by the external generator (i.e., Qwen1.5-32B-Chat) for fine-tuning the policy.}
    \label{tab:ablation}
    \vspace{-0.45cm}
\end{wraptable}
progressive strategy of internal and external evolution. As shown in Table \ref{tab:ablation}, directly fine-tuning the policy with \emph{chosen} responses (SFT-Chosen) from the UltraFeedback dataset improves the score from 6.33 to 6.99. Utilizing responses generated by Qwen1.5-32B-Chat (SFT-External Generator) leads to a slightly higher score of 7.06. These results underscore the benefits of using high-quality annotated or generated data. Furthermore, revising the responses from the external generator with our adaptive reviser (External Evolution) results in an even higher score of 7.11, highlighting the critical role of adaptive refinement in enhancing model performance.

Interestingly, fine-tuning with internal evolution yields a score of 6.89, which is less effective compared to the direct use of high-quality generated responses (SFT-External Generator). However, the most significant improvement is observed when combining internal and external evolution, achieving a score of 7.32. This combined strategy effectively leverages initial improvements from the policy model’s revised responses, followed by further enhancements from an external generator’s refined responses. Moreover, the progressive policy optimization strategy demonstrates superior performance over individual internal or external strategies when various amounts of additional data (30K, 60K) are utilized, demonstrating the robustness and scalability of our SEFT framework.

\section{Conclusion}
In this paper, we introduce self-evolution fine-tuning (SEFT) for alignment of large language models (LLMs). SEFT employs an adaptive reviser to enhance the overall quality of initial responses by making revisions only when improvements are feasible. Our experiments show that the adaptive reviser consistently enhances response quality and outperforms baseline methods. The effectiveness of SEFT in policy optimization is validated through extensive experiments on benchmarks such as AlpacaEval 2.0 and MT-Bench. Notably, we highlight SEFT’s capability to leverage unannotated data to improve response quality while maintaining stability and efficiency in policy optimization.

\bibliographystyle{plainnat}
\bibliography{neurips_2024}

\clearpage
\appendix

\section{Performance of Adaptive Reviser}
\label{reviser_performence}
\begin{figure}[h]
\vspace{-0.3cm}
  \begin{center}
 \includegraphics[width=1.00\textwidth]{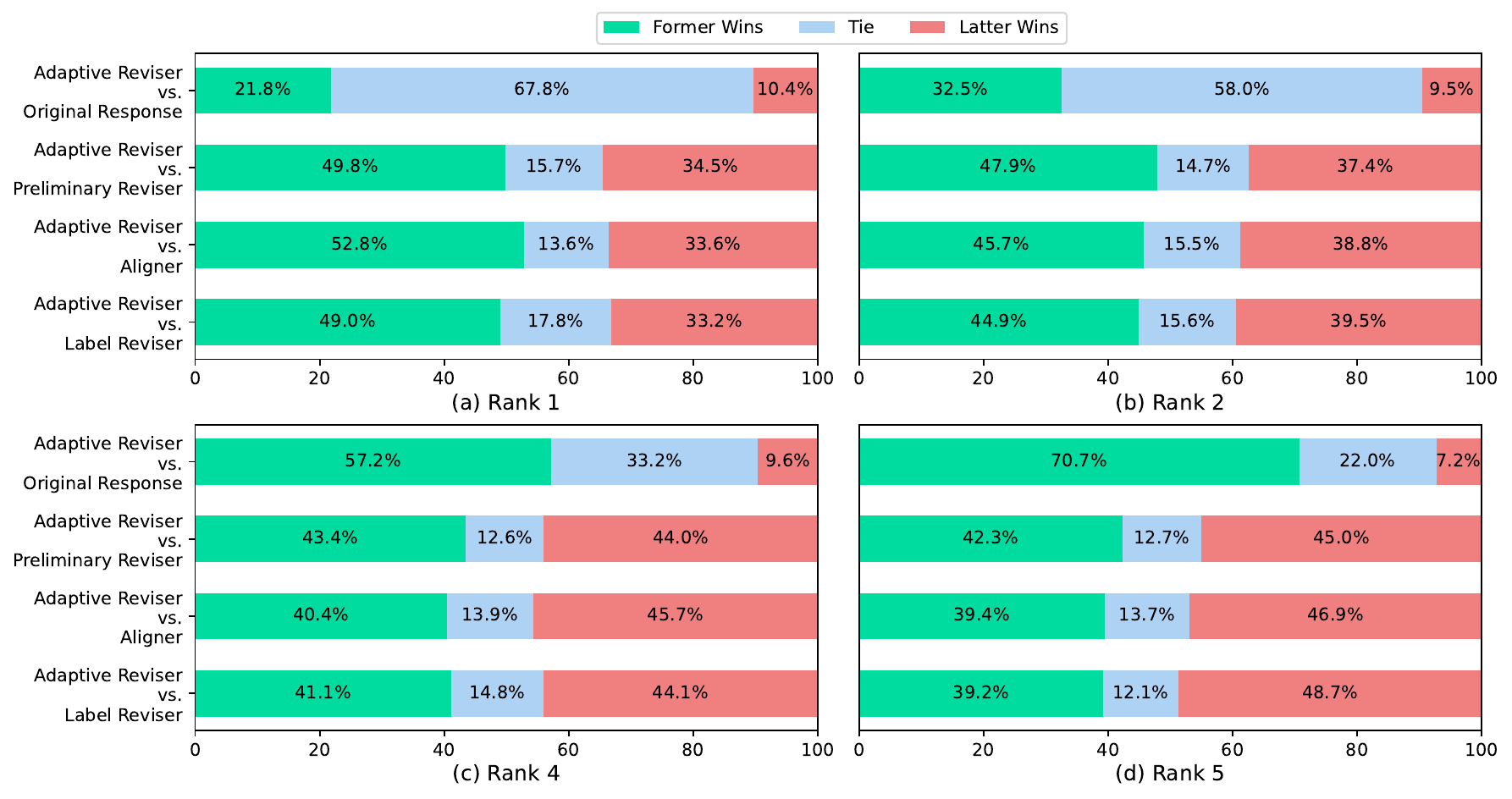}
  \end{center}
  \caption{Performance comparison of the adaptive reviser and baseline methods on the Nectar test
set. Subfigures (a), (b), (c), and (d) illustrate the original responses are at rank 1, 2, 4, and 5, respectively. }
  \label{fig:pre_reviser_compare_appendix}
\end{figure}

Figure \ref{fig:pre_reviser_compare_appendix} shows the performance comparison of the adaptive reviser and baseline methods on the Nectar test set for responses ranked 1, 2, 4, and 5.

\begin{table*}[h]
\centering
\resizebox{0.999\textwidth}{!}{
\begin{threeparttable}
\begin{tabular}{lcccccccc}
    \toprule
     \multirow{2}{*}{\textbf{Reviser Backbone}}& \multirow{2}{*}{\textbf{Original Response}}&  \multicolumn{4}{c}{\textbf{Benefit Score}} & \multicolumn{3}{c}{\textbf{Proportion of Revision Label}}\\
     \cmidrule(lr){3-6}
     \cmidrule(lr){7-9}
     ~ & ~ & RM1 & RM2 & RM3 & RM4  & No Revise & Major Revise & Minor Revise \\
    \midrule
    \multirow{8}{*}{OpenChat-3.5-7B}
& Rank 0  & +0.07 & +9.64 & +0.09 & \textbf{+0.10} & 75.72\% & 15.18\% & 9.11\% \\
& Rank 1 & +0.15 & +19.58 & +0.16 & +0.23 & 65.87\% & 22.74\% & 11.40\% \\
& Rank 2  & +0.33 & +38.03 & +0.33 & +0.48 & 54.95\% & 32.32\% & 12.73\% \\
& Rank 3  & +0.61 & +64.28 & +0.62 & +0.78 & 43.66\% & 43.61\% & 12.73\% \\
& Rank 4 & +1.07 & +107.05 & +1.08 & +1.26 & 29.77\% & 58.68\% & 11.55\% \\
& Rank 5  & +2.06 & +207.61 & +2.03 & +2.26 & 18.74\% & 71.78\% & 9.37\% \\
& Rank 6  & +3.20 & +314.23 & +3.09 & +3.28 & 9.74\% & 85.30\% & 4.95\% \\
\rowcolor[gray]{.9} \cellcolor{white} & Averages & +1.07  & +108.63  & +1.06  & +1.20  & 42.64\% & 47.09\% & 10.26\% \\

     \midrule
    \multirow{8}{*}{Qwen1.5-32B-Chat}
& Rank 0  & +0.18 & \textbf{+18.20} & +0.17 & +0.09 & 83.33\% & 14.06\% & 2.61\% \\
& Rank 1 & +0.30 & \textbf{+30.57} & +0.27 & +0.25 & 74.71\% & 22.79\% & 2.50\% \\
& Rank 2  & +0.54 & +53.92 & +0.48 & +0.52 & 63.79\% & 33.87\% & 2.29\% \\
& Rank 3  & +0.87 & +87.00 & +0.82 & +0.86 & 50.21\% & 46.96\% & 2.82\% \\
& Rank 4 & +1.51 & +145.62 & +1.43 & +1.42 & 36.16\% & 61.40\% & 2.45\% \\
& Rank 5  & \textbf{+2.63} & \textbf{+255.75} & \textbf{+2.49} & \textbf{+2.48} & 24.44\% & 74.44\% & 1.12\% \\
& Rank 6  & \textbf{+4.01} & \textbf{+383.56} & \textbf{+3.76} & \textbf{+3.62} & 11.66\% & 87.49\% & 0.85\% \\
\rowcolor[gray]{.9} \cellcolor{white} & Average & \textbf{+1.43}  & \textbf{+139.23}  & \textbf{+1.35}  & +1.32  & 49.19\% & 48.72\% & 2.09\% \\

     \midrule
     \multirow{8}{*}{Yi-34B-Chat}
& Rank 0  & \textbf{+0.19} & +17.92 & \textbf{+0.18} & \textbf{+0.10} & 75.40\% & 13.26\% & 11.34\% \\
& Rank 1 & \textbf{+0.32} & +30.07 & \textbf{+0.30} & \textbf{+0.27} & 65.87\% & 21.30\% & 12.83\% \\
& Rank 2  & \textbf{+0.55} & \textbf{+55.96} & \textbf{+0.50} & \textbf{+0.55} & 52.82\% & 32.11\% & 15.07\% \\
& Rank 3  & \textbf{+0.90} & \textbf{+90.93} & \textbf{+0.86} &\textbf{ +0.91} & 40.58\% & 43.56\% & 15.87\% \\
& Rank 4 & \textbf{+1.52} & \textbf{+146.53} & \textbf{+1.46} & \textbf{+1.43} & 27.00\% & 59.42\% & 13.58\% \\
& Rank 5  & +2.59 & +254.44 & \textbf{+2.49} & \textbf{+2.48} & 16.67\% & 73.43\% & 9.90\% \\
& Rank 6  & +3.88 & +374.36 & +3.67 & +3.61 & 8.52\% & 86.42\% & 5.06\% \\
\rowcolor[gray]{.9} \cellcolor{white} & Average & +1.42  & +138.60  & \textbf{+1.35}  & \textbf{+1.34}  & 40.98\% & 47.07\% & 11.95\% \\

\bottomrule
\end{tabular}
\end{threeparttable}
}
\caption{Performance of the adaptive reviser with various backbones on the Nectar test set. We used four different reward models (RMs) for scoring: RM1-RM4, specifically FsfairX-LLaMA3-RM-v0.1, Eurus-RM-7b, RM-Mistral-7B, and Starling-RM-7B-alpha. The highest score for each response is highlighted in bold. The columns under ``Benefit Score'' reflect the improvements of the revised responses compared to the original responses. Meanwhile, the columns under ``Proportion of Revision Label'' show the distribution of different revision labels applied to the original responses.}
\label{tab:main_reviser_results}
\end{table*}

\begin{table*}[t!]
\centering
\resizebox{0.999\textwidth}{!}{
\begin{threeparttable}
\begin{tabular}{lcccccccccc}
    \toprule
     \multirow{2}{*}{\textbf{Reviser}}& \multirow{2}{*}{\textbf{Original Response}} &  \multicolumn{2}{c}{\textbf{RM1}} & \multicolumn{2}{c}{\textbf{RM2}} & \multicolumn{2}{c}{\textbf{RM3}} & \multicolumn{2}{c}{\textbf{RM4}} \\
     \cmidrule(lr){3-4} \cmidrule(lr){5-6} \cmidrule(lr){7-8} \cmidrule(lr){9-10}
      ~ & ~ & Score & Benefit &  Score &  Benefit &  Score &  Benefit &  Score & Benefit \\
    \midrule
    \multirow{8}{*}{Original Response}
& Rank 0 & -1.52 & - & 64.81 & - & 5.08 & - & 3.42 & - \\
& Rank 1 & -1.85 & - & 35.98 & - & 4.79 & - & 3.04 & - \\
& Rank 2 & -2.24 & - & -6.08 & - & 4.42 & - & 2.61 & - \\
& Rank 3 & -2.73 & - & -50.74 & - & 3.94 & - & 2.16 & - \\
& Rank 4 & -3.41 & - & -110.68 & - & 3.30 & - & 1.59 & - \\
& Rank 5 & -4.60 & - & -227.37 & - & 2.18 & - & 0.50 & - \\
& Rank 6 & -5.95 & - & -351.74 & - & 0.93 & - & -0.64 & - \\
\rowcolor[gray]{.9} \cellcolor{white} & Average & -3.19  & - & -92.26  & - & 3.52  & - & 1.81  & - \\ 

     \midrule
    \multirow{8}{*}{Adaptive Reviser}
& Rank 0 & \textbf{-1.45} & \textbf{+0.07} & \textbf{74.44} & \textbf{+9.64} & \textbf{5.17} & \textbf{+0.09} & \textbf{3.51} & \textbf{+0.10} \\
& Rank 1 & \textbf{-1.70} & \textbf{+0.15} & \textbf{55.55} & \textbf{+19.58} & \textbf{4.95} & \textbf{+0.16} & \textbf{3.27} & \textbf{+0.23} \\
& Rank 2 & \textbf{-1.92} & \textbf{+0.33} & \textbf{31.95} & \textbf{+38.03} & \textbf{4.75} & \textbf{+0.33} & \textbf{3.08} & \textbf{+0.48} \\
& Rank 3 & \textbf{-2.12} & \textbf{+0.61} & 13.55 & +64.28 & 4.56 & +0.62 & 2.94 & +0.78 \\
& Rank 4 & -2.35 & +1.07 & -3.63 & +107.05 & 4.38 & +1.08 & 2.85 & +1.26 \\
& Rank 5 & -2.54 & +2.06 & -19.76 & +207.61 & 4.22 & +2.03 & 2.76 & +2.26 \\
& Rank 6 & -2.75 & +3.20 & -37.50 & +314.23 & 4.02 & +3.09 & 2.64 & +3.28 \\
\rowcolor[gray]{.9} \cellcolor{white} & Average & \textbf{-2.12} & \textbf{+1.07} & \textbf{16.37} & \textbf{+108.63} & \textbf{4.58} & \textbf{+1.06} & \textbf{3.01} & \textbf{+1.20} \\

     \midrule
    \multirow{8}{*}{Preliminary Reviser}
& Rank 0 & -2.04 & -0.52 & 29.33 & -35.48 & 4.68 & -0.39 & 3.09 & -0.33 \\
& Rank 1 & -2.11 & -0.26 & 24.15 & -11.82 & 4.61 & -0.18 & 3.05 & +0.01 \\
& Rank 2 & -2.17 & +0.07 & 15.62 & +21.70 & 4.54 & +0.12 & 2.98 & +0.37 \\
& Rank 3 & -2.26 & +0.47 & 8.54 & +59.27 & 4.48 & +0.53 & 2.93 & +0.77 \\
& Rank 4 & -2.41 & +1.00 & -1.59 & +109.09 & 4.35 & +1.05 & 2.88 & +1.29 \\
& Rank 5 & -2.52 & +2.08 & -9.62 & +217.75 & 4.26 & +2.08 & 2.83 & +2.32 \\
& Rank 6 & -2.66 & +3.28 & -25.45 & +326.29 & 4.11 & +3.19 & 2.73 & +3.36 \\
\rowcolor[gray]{.9} \cellcolor{white} & Average & -2.31 & +0.88 & 5.85 & +98.11 & 4.43 & +0.91 & 2.93 & +1.11 \\

     \midrule
    \multirow{8}{*}{Aligner}
& Rank 0 & -2.04 & -0.52 & 27.18 & -37.63 & 4.66 & -0.42 & 3.09 & -0.33 \\
& Rank 1 & -2.10 & -0.25 & 17.71 & -18.26 & 4.60 & -0.19 & 3.02 & -0.02 \\
& Rank 2 & -2.10 & +0.15 & 16.89 & +22.96 & 4.58 & +0.16 & 3.02 & +0.41 \\
& Rank 3 & -2.14 & +0.59 & \textbf{15.71} & \textbf{+66.45} & \textbf{4.57} & \textbf{+0.63} & 2.98 & +0.82 \\
& Rank 4 & \textbf{-2.28} & \textbf{+1.14} & \textbf{6.44} & \textbf{+117.12} & \textbf{4.46} & \textbf{+1.16} & \textbf{2.92} & \ \textbf{+1.33} \\
& Rank 5 & \textbf{-2.40} & \textbf{+2.20} & \textbf{-5.00} & \textbf{+222.36} & \textbf{4.37} & \textbf{+2.18} & 2.86 & +2.36 \\
& Rank 6 & \textbf{-2.48} & \textbf{+3.47} & \textbf{-12.04} & \textbf{+339.70} & \textbf{4.29} & \textbf{+3.36} & \textbf{2.82} & \textbf{+3.46} \\
\rowcolor[gray]{.9} \cellcolor{white} & Average & -2.22 & +0.97 & 9.55 & +101.81 & 4.50 & +0.98 & 2.96 & +1.15 \\

     \midrule
    \multirow{8}{*}{Label Reviser}
& Rank 0 & -2.03 & -0.52 & 29.38 & -35.42 & 4.68 & -0.40 & 3.16 & -0.26 \\
& Rank 1 & -2.10 & -0.24 & 24.53 & -11.44 & 4.61 & -0.18 & 3.06 & +0.02 \\
& Rank 2 & -2.11 & +0.14 & 20.29 & +26.37 & 4.59 & +0.17 & 3.03 & +0.43 \\
& Rank 3 & -2.20 & +0.53 & 10.30 & +61.04 & 4.49 & +0.55 & \textbf{2.99} & \textbf{+0.83} \\
& Rank 4 & -2.33 & +1.08 & 1.09 & +111.77 & 4.42 & +1.11 & \textbf{2.92} & \textbf{+1.33} \\
& Rank 5 & -2.41 & +2.19 & -5.82 & +221.55 & 4.35 & +2.17 & \textbf{2.89} & \textbf{+2.39} \\
& Rank 6 & -2.55 & +3.40 & -18.12 & +333.61 & 4.24 & +3.32 & 2.79 & +3.43 \\
\rowcolor[gray]{.9} \cellcolor{white} & Average & -2.25 & +0.77 & 8.81 & +85.06 & 4.48 & +0.80 & 2.98 & +1.02 \\

\bottomrule
\end{tabular}
\end{threeparttable}
}
\caption{Performance comparison between our adaptive reviser and baseline methods, with Openchat-3.5-7B serving as the backbone where applicable. The highest score for each response is highlighted in bold. The specifications of RM1-RM4 are the same as in Table \ref{tab:main_reviser_results}. The results show that our adaptive reviser consistently outperforms the baseline methods, especially for the top-ranked responses.}

\label{tab:pre_reviser_results}
\vspace{-0.3cm}
\end{table*}

We trained the adaptive revisers of three different sizes based on various backbones and tested their performance on our Nectar test set. The benefit score is defined as: \( s_i = \texttt{Score} (\hat{Y}_i^l) - \texttt{Score} (Y_i^l)\), where \(\texttt{Score} (Y_i^l)\) and \(\texttt{Score} (\hat{Y}_i^l)\) are the reward scores of the original response and revised response evaluated by reward model, respectively. We utilized the four reward models mentioned above to obtain the benefit scores. Additionally, we analyzed the distribution of revision labels when revisers corrected the original responses.

As shown in Table \ref{tab:main_reviser_results}, all revisers exhibit consistently superior performance when handling responses of varying quality. Specifically, when the original response quality is low, the reviser’s improvement yields a significantly high benefit. For instance, the reviser based on Qwen1.5-32B-Chat achieves benefit scores of +4.01, +383.56, +3.76, and +3.62 across the four RMs when revising Rank 6 responses. Conversely, when the initial response quality is high (GPT-4 level), the reviser still manages to bring about stable quality enhancements. Furthermore, we observed that even the 7B reviser (OpenChat-3.5-7B) could improve high-quality original responses, despite the original responses being generated by models larger than 7B.
The distribution of revision labels indicates that as the original response quality increases, the proportion of \texttt{[No Revise]} labels output by the reviser also increases. This aligns with our primary goal for implementing revision labels: to ensure that revisions are made only when necessary, preserving high-quality responses as they are. On the other hand, as the quality of the initial response decreases, the proportion of \texttt{[Major Revise]} labels increases, signaling a more extensive revision effort by the reviser. The \texttt{[Minor Revise]} label appears more frequently for responses of moderate quality. This is expected because the \texttt{[Minor Revise]} label acts as a middle ground between \texttt{[No Revise]} and \texttt{[Major Revise]}.

The experimental results in Table \ref{tab:pre_reviser_results} highlight the exceptional performance of the adaptive reviser compared to baseline methods. Across all reward models (RM1-RM4), the adaptive reviser achieves the highest average benefit scores, indicating its overall superiority. Notably, the adaptive reviser shows significant improvements over the original responses, particularly for lower-quality inputs, with greater benefit scores as the initial response quality declines. Furthermore, for high-quality original responses, the adaptive reviser consistently outperforms other revisers. For instance, under RM2, the adaptive reviser achieves benefit scores of +9.64 and +19.58 for Rank 0 and 1, respectively, while other methods show considerably lower or even negative scores. These results demonstrate that the adaptive reviser excels in refining both high and low-quality responses.
\begin{table*}[t!]
    \centering
    \resizebox{0.99\textwidth}{!}{
    \begin{threeparttable}
    
    \begin{tabular}{cccc}
    \toprule
        \multirow{2}{*}{Hyperparameters} & Preliminary & Adaptive & Internal (External)\\
        ~ & Reviser & Reviser & Evolution \\
        \midrule
        epochs & 3 & 3 & 3 \\
        max-length & 2048 & 2048 & 2048 \\
        per-device-train-batch-size & 2 & 2 & 4 \\
        gradient-accumulation-steps & 8 & 8 & 4 \\
        learning-rate & 2e-5 & 2e-5 & 5e-7 \\
        LR-scheduler-type & cosine & cosine & linear \\
        LR-warmup-ratio & 0.03 & 0.03 & 0.03 \\
        weight-decay & 0.0 & 0.0 & 0.0 \\
        gradient-checkpointing & TRUE & TRUE & TRUE \\
        seed & 42 & 42 & 42 \\
        zero-stage & 3 & 3 & 3 \\
        optimizer & AdamW & AdamW & AdamW \\
        bf16 & TRUE & TRUE & TRUE \\
        tf32 & TRUE & TRUE & TRUE \\
        dataset-size & 53K & 124K & 61K, 91K, 121K \\
        \bottomrule
    \end{tabular}
    \end{threeparttable}
    }
    \caption{Hyperparameters of our training in different stages.}
    \label{tab:hyperp}
\end{table*}

\section{Implementation Details}
\label{appendix:implementation}
The hyperparameters for our implementations are listed in Table \ref{tab:hyperp}. For the label reviser, we utilize three revision labels: \texttt{[No Revise]}, \texttt{[Major Revise]}, and \texttt{[Minor Revise]}. Specifically, 10\% of the Nectar training samples with original responses ranked at 0 are labeled as \texttt{[No Revise]}.  For the remaining samples, those with original response rankings between 1 and 3 receive the \texttt{[Minor Revise]} label, while all other samples are assigned the \texttt{[Major Revise]} label.

\section{Case Study}
In this subsection, we provide examples of prompts and original responses from the Nectar training set. We illustrate how revision labels are assigned by comparing the scores given by the Starling RM to the original responses and those revised by our preliminary reviser. This comparison highlights the differences in response quality that guide the assignment of \texttt{[No Revise]}, \texttt{[Minor Revise]}, and \texttt{[Major Revise]} labels for training the adaptive reviser.

\begin{table}[htbp]
    \centering
    \resizebox{0.99\textwidth}{!}{
    \renewcommand{\arraystretch}{1.5}
    \begin{tabular}{lp{10cm}}
        \toprule
        \textbf{Prompt} & Human: Been listening to a lot of Aretha Franklin lately and really enjoying it. Do you know any female singing artists that are like her?\textbackslash n \textbackslash n Assistant: I’ve heard people mention Nina Simone, Ella Fitzgerald, and Patsy Cline.\textbackslash n \textbackslash n Human: Nina and Ella seem like good suggestions. Isn't Patsy Cline a bit country and western though?\textbackslash n \textbackslash n Assistant:  \\ 
        \midrule
        \textbf{Original Response} & She was known as a Country singer, but she also did some more traditional standards. \\ 
        \midrule
        \textbf{Revised Response} & Yes, you're correct. Patsy Cline is often associated with country and western music. However, she did have a powerful and soulful voice, which might be similar to Aretha Franklin's in terms of emotional depth. If you're specifically looking for female singers with a soul and R\&B style similar to Aretha Franklin, you might also enjoy artists like Etta James, Dionne Warwick, and Gladys Knight. \\ 
        \midrule
        \textbf{Scored by Starling RM} & Original Response: -0.5859375; Revised Response: 5.84375 \\ 
        \midrule
        \textbf{Revision Label} & \texttt{[Major Revise]} \\ 
        \bottomrule
        \\
    \end{tabular}
    }
    \caption{A prompt and the original response from the Nectar test set, along with a revised response by our preliminary reviser. The revised response receives a significantly higher score (\( s_i \) > 1.0) than the original response, as evaluated by Starling RM. This substantial improvement demonstrates that the original response was easy for the preliminary reviser to enhance significantly. As a result, the revision label for this instance is assigned as \texttt{[Major Revise]}, demanding a substantial revision.}

    \label{tab:my_label1}
\end{table}

\begin{table}[htbp]
    \centering
    \resizebox{0.99\textwidth}{!}{
    \renewcommand{\arraystretch}{1.5}
    \begin{tabular}{lp{10cm}}
        \toprule
        \textbf{Prompt} & Human: How do I tell my kid not to lie? \textbackslash n \textbackslash n Assistant: \\ \midrule
        
        \textbf{Original Response} & Here are a few tips for talking to your child about lying: \textbackslash n  \textbackslash n 1. Explain the consequences: Let your child know that lying can have serious consequences, such as losing trust and damaging relationships. \textbackslash n  \textbackslash n 2. Be a role model: Children often learn by example, so make sure you are setting a good example by being honest and truthful yourself. \textbackslash n  \textbackslash n 3. Encourage honesty: Praise your child when they tell the truth, even if it's something you don't want to hear. This will reinforce the importance of honesty. \textbackslash n  \textbackslash n 4. Talk about the difference between truth and lies: Help your child understand that there is a difference between telling the truth and telling a lie. Explain that telling the truth is always the right thing to do. \textbackslash n  \textbackslash n 5. Discuss the reasons behind lying: Sometimes children lie because they are afraid of getting in trouble or they want to avoid disappointing someone. Talk to your child about these reasons and help them find alternative ways to handle these situations. \textbackslash n  \textbackslash n 6. Set clear expectations: Let your child know that lying is not acceptable and that there will be consequences if they are caught lying. \textbackslash n  \textbackslash n 7. Listen to your child: If your child does lie, try to understand why they did it. Listen to their side of the story and use it as an opportunity to teach them about honesty and trust. \textbackslash n  \textbackslash n Remember to be patient and consistent when talking to your child about lying. It may take time for them to understand the importance of honesty, but with your guidance, they will learn to value the truth. \\ 
        \midrule
        
        \textbf{Revised Response} & 1. Lead by example: Children learn by observing their parents and caregivers. Make sure you are honest and truthful in your own actions and words. \textbackslash n  \textbackslash n 2. Explain the consequences: Talk to your child about the negative consequences of lying, such as losing trust and damaging relationships. \textbackslash n  \textbackslash n 3. Encourage open communication: Create a safe and open environment for your child to talk to you about anything without fear of punishment or judgment. \textbackslash n  \textbackslash n 4. Teach the difference between right and wrong: Help your child understand the difference between telling the truth and telling a lie. \textbackslash n  \textbackslash n 5. Praise honesty: When your child tells the truth, acknowledge and praise their honesty to reinforce positive behavior. \textbackslash n  \textbackslash n 6. Address the root cause: If your child is lying, try to understand the reason behind it. Are they afraid of getting in trouble? Are they trying to impress others? Addressing the underlying issue can help prevent future lying. \textbackslash n  \textbackslash n 7. Set clear expectations: Make it clear that lying is not acceptable and there will be consequences for doing so. \textbackslash n  \textbackslash n 8. Use age-appropriate examples: Use stories or examples that your child can relate to in order to illustrate the importance of honesty. \textbackslash n  \textbackslash n 9. Be a good listener: Sometimes, children lie because they feel unheard or ignored. Make sure to actively listen to your child and give them a chance to express themselves. \textbackslash n  \textbackslash n 10. Seek professional help if needed: If your child continues to struggle with lying despite your efforts, it may be helpful to seek guidance from a child psychologist or therapist. \\ 
        \midrule
        \textbf{Scored by Starling RM} & Original Response: 3.453125;  Revised Response: 3.984375 \\ 
        \midrule
        \textbf{Revision Label} & \texttt{[Minor Revise]} \\ 
        \bottomrule
        \\
    \end{tabular}
    }
    \caption{A prompt and the original response from the Nectar test set, along with a revised response by our preliminary reviser. The revised response receives a slightly higher score (1.0 > \( s_i \) = 0.53125 > 0.3) than the original response, as evaluated by Starling RM. This improvement indicates that the preliminary reviser can enhance the original response, although with a little bit of difficulty. Therefore, with a probability of 0.8, the revision label is assigned as \texttt{[Minor Revise]}, signifying a minor revision.}

    \label{tab:my_label2}
\end{table}

\begin{table}[htbp]
    \centering
    \resizebox{0.99\textwidth}{!}{
    \renewcommand{\arraystretch}{1.5}
    \begin{tabular}{lp{10cm}}
        \toprule
        \textbf{Prompt} & Human: What's a Greek mythology that is about evil parents creating a good child (like NAME\_1?)\textbackslash n \textbackslash n Assistant:  \\ 
        \midrule
        
        \textbf{Original Response} & There are several Greek myths involving evil parents creating a good child. Here are a few examples: \textbackslash n  \textbackslash n 1. The story of Persephone: In this myth, Hades, the god of the underworld, falls in love with Persephone, the daughter of Demeter, the goddess of agriculture. Hades kidnaps Persephone and takes her to the underworld, where she becomes his queen. Demeter is grief-stricken and refuses to let the earth produce any crops until her daughter is returned. Eventually, a deal is made between Hades and Demeter that allows Persephone to return to the world above, but she must spend part of each year in the underworld with Hades, which is why the seasons change. \textbackslash n 2. The story of Dionysus: In this myth, the goddess Semele, who is the mother of Dionysus, is killed by her husband, King Cadmus, after he becomes jealous of her affair with Zeus. Zeus, who is the father of Dionysus, saves his son by placing him in his thigh, where he grows to maturity. Dionysus is known for his role in promoting fertility and the growth of crops, and his birth is seen as a symbol of the cycle of life and death. \textbackslash n 3. The story of Hercules: In this myth, Hercules' parents, Zeus and Alcmene, are not evil, but Hercules' stepfather, Hera, is often depicted as a villainous figure. Hera is jealous of Hercules' divine parentage and seeks to harm him throughout his life. Despite this, Hercules is known for his heroism and his role in protecting the world from various monsters and evils. \textbackslash n These are just a few examples of Greek myths involving evil parents creating a good child. There are many other myths that explore similar themes, and the specific details of each myth can vary depending on the source and interpretation. \\ 
        \midrule
        
        \textbf{Revised Response} & One Greek myth that fits this description is the story of Persephone. In this myth, Persephone was the daughter of Demeter, the goddess of agriculture, and Zeus, the king of the gods. Despite her divine heritage, Persephone was kidnapped by Hades, the god of the underworld, and forced to become his wife. However, Persephone was able to maintain her innocence and purity, and she eventually became a beloved figure in Greek mythology known for her kindness and compassion. \\ 
        \midrule
        \textbf{Scored by Starling RM} & Original Response: 4.25;  Revised Response: 2.1875 \\ 
        \midrule
        \textbf{Revision Label} & \texttt{[No Revise]} \\
        \bottomrule
        \\
    \end{tabular}
    }
    \caption{A prompt and the original response from the Nectar test set, along with a revised response by our preliminary reviser. The revised response receives a much lower score (\( s_i \) < 0.3) than the original response, as evaluated by Starling RM. This 
 means that the original response is of high quality and is difficult for the preliminary reviser to enhance. Therefore, with a probability of 0.8, the
revision label for this instance is assigned as \texttt{[No Revise]}, indicating that no revision is needed.}

    \label{tab:my_label3}
\end{table}

\end{document}